\documentclass[runningheads]{llncs}
\usepackage[T1]{fontenc}
\usepackage{graphicx,verbatim}
\usepackage[misc]{ifsym}
\usepackage{amsmath}
\usepackage{booktabs}
\usepackage{amssymb}
\usepackage{multirow}
\usepackage{tabularx}
\usepackage[marginal]{footmisc}

\begin{document}
\title{Progressive Self-Supervised Learning with Individualized Community Assignment for Brain Network Analysis}
\titlerunning{BrainPICM}
%\titlerunning{Abbreviated paper title}
% If the paper title is too long for the running head, you can set
% an abbreviated paper title here
%
\begin{comment}  %% Removed for anonymized MICCAI submission
\author{First Author\inst{1}\orcidID{0000-1111-2222-3333} \and
Second Author\inst{2,3}\orcidID{1111-2222-3333-4444} \and
Third Author\inst{3}\orcidID{2222--3333-4444-5555}}
%
\authorrunning{F. Author et al.}
% First names are abbreviated in the running head.
% If there are more than two authors, 'et al.' is used.
%
\institute{Princeton University, Princeton NJ 08544, USA \and
Springer Heidelberg, Tiergartenstr. 17, 69121 Heidelberg, Germany
\email{lncs@springer.com}\\
\url{http://www.springer.com/gp/computer-science/lncs} \and
ABC Institute, Rupert-Karls-University Heidelberg, Heidelberg, Germany\\
\email{\{abc,lncs\}@uni-heidelberg.de}}

\end{comment}

\author{Hairui Chen\inst{1,2} %index{Yang, Yanwu}
\and
Yanwu Yang\inst{3,4} %index{Chen, Hairui}
\and
Jianfeng Cao\inst{1} %index{Hu, Jiesi}
\and
Hanyang Peng \inst{2} %index{Guo, Xutao}
\and
Chenfei Ye \inst{1}
\and
Ting Ma \inst{1}$^{(\textrm{\Letter})}$ %index{Ma, Ting}
}  %% Added for anonymized MICCAI submission
\authorrunning{H. Chen et al.}
\institute{
Harbin Institute of Technology at Shenzhen, Shenzhen, China\\ \email{tma@hit.edu.cn}
\\
 \and
Peng Cheng Laboratory, Shenzhen, China \\
\and 
University of Tübingen, Tübingen, Germany \\
\and
Max Planck Institute for Intelligent Systems, Tübingen, Germany
}
  
\maketitle              % typeset the header of the contribution
\footnote{H. Chen and Y. Yang contributed equally to this work.}
\begin{abstract}
Brain networks exhibit a modular community structure that varies across individuals and neurological conditions. However, existing self-supervised learning (SSL) methods often overlook this heterogeneity, relying on generic masking strategies that fail to capture subject-specific functional organization. We propose BrainPICM, a self-supervised framework for brain network analysis via progressive individualized community-aware masking. BrainPICM formulates ROI-to-community mapping as a progressive unbalanced optimal transport process, yielding soft assignments and per-ROI confidence scores. Guided by these confidence estimates, a curriculum-style masking strategy gradually incorporates low-confidence, potentially pathological regions into training, enabling the model to learn both stable modular structures and individual variations. Additionally, a deviation-aware aggregation module quantifies functional reorganization by measuring mass redistribution relative to a population template, enhancing interpretability and downstream prediction. Experiments on three fMRI datasets (ABIDE-I, ADHD-200, ADNI) show that BrainPICM consistently outperforms state-of-the-art supervised and SSL methods in diagnostic accuracy, indicating that explicitly injecting modular community structure into masked modeling yields more functionally consistent and generalizable representations. The source code for this approach will be released at \url{https://github.com/Hrychen7/BrainPICM}.

\keywords{Self-supervised Learning  \and Brain Network Analysis \and Individualized Community Assignment.}
% Authors must provide keywords and are not allowed to remove this Keyword section.

\end{abstract}
\section{Introduction}
% Brain networks have become a pivotal paradigm in cognitive neuroscience, enabling the characterization of complex relationships between brain abnormalities and behavioral changes through network-based modeling \cite{yang2023deep}. Functional magnetic resonance imaging (fMRI) measures blood-oxygen-level-dependent (BOLD) signals to investigate the functional organization of brain regions and their interactions. By estimating correlations between BOLD signals across regions of interest (ROIs), fMRI constructs functional brain networks and provides insights into the neural mechanisms underlying neurodevelopmental disorders \cite{lei2024hybrid}.
Brain networks have become a pivotal paradigm in cognitive neuroscience, enabling the characterization of complex relationships between brain abnormalities and behavioral changes through network-based modeling \cite{yang2023deep}. Functional magnetic resonance imaging (fMRI) measures blood-oxygen-level-dependent (BOLD) signals to investigate the functional organization of brain regions and their interactions \cite{yang2025hierarchical}. A fundamental property of functional brain networks is their modular community organization, which supports coordinated information processing~\cite{sporns2016modular,yeo2011organization}. Importantly, these communities vary across subjects and may be reorganized in disease states, reflecting subject-specific functional heterogeneity~\cite{keller2023personalized,dujardin2020tau,segal2023regional}.

Learning robust representations from functional connectivity is challenging. In addition to the modular community organization that varies across individuals, brain networks are high-dimensional, training samples are often limited, and substantial inter-subject variability exists across cohorts and disorders~\cite{yang2024advancing}. These factors can make purely supervised models prone to overfitting and less reliable generalization. Compared with purely supervised learning, self-supervised learning (SSL) provides a promising alternative by exploiting intrinsic structure in unlabeled networks, which can improve generalization while reducing reliance on large annotated datasets.
% Recently, advanced artificial intelligence technologies such as supervised deep learning have emerged as a powerful tool to improve the disease diagnosis performance and understand neural mechanisms\cite{kan2022brain,pei2025community}.
% Nevertheless, several challenges remain in brain imaging analysis using supervised deep learning approaches. First, while brain imaging  are inherently high-dimensional, the deep learning models are heavily constrained by the limited amount of data samples\cite{yang2024advancing}. Moreover, brain imaging exhibits heterogeneity in etiology and phenotypic for the brain disorders\cite{dujardin2020tau,segal2023regional}, which may result in suboptimal performance due to overfitting and potential biases.

Among SSL paradigms, masked modeling has recently gained popularity for brain network representation learning, where models are trained to reconstruct functional connectivity from masked ROIs. Existing approaches mainly differ in masking policies, ranging from uniform random masking~\cite{jung2023eag,yang2024brainmass} to attention-guided strategies~\cite{zhao2024tardrl}. While effective in learning generic features, these masking policies may not explicitly incorporate functional community structure. Consequently, the learned representations may be less functionally consistent and may generalize less reliably when community organization varies across individuals.

To address this gap, we propose BrainPICM, a self-supervised framework for \textbf{Brain} network analysis using \textbf{P}rogressive \textbf{I}ndividualized \textbf{C}ommunity-aware \textbf{M}asking to learn personalized yet consistent representations. Specifically, BrainPICM models ROI-to-community mapping via a progressive unbalanced optimal transport process, yielding soft subject-specific assignments and confidence scores while robustly handling outlier regions. Guided by these scores, a curriculum-inspired masking strategy gradually introduces low-confidence, potentially pathological ROIs into training to capture individual heterogeneity. Additionally, we summarize the soft assignments into compact community-level deviation features and fuse them with ROI-level connectivity features for downstream diagnosis. Our contributions are summarized below:

(1)~We propose BrainPICM, a self-supervised framework with community-guided curriculum masking driven by soft ROI-to-community assignments and ROI confidence.

(2)~We formulate individualized soft community assignment as a progressive unbalanced optimal transport (UOT) process with a virtual community, yielding $Q$ and ROI-level confidence $q$ under functional heterogeneity.

(3)~We derive compact community-level deviation features from $Q$ and fuse them with ROI-level connectivity features in a lightweight dual-branch predictor for improved diagnosis.

\begin{figure*}[t]
\centering
\includegraphics[width=1.0\textwidth]{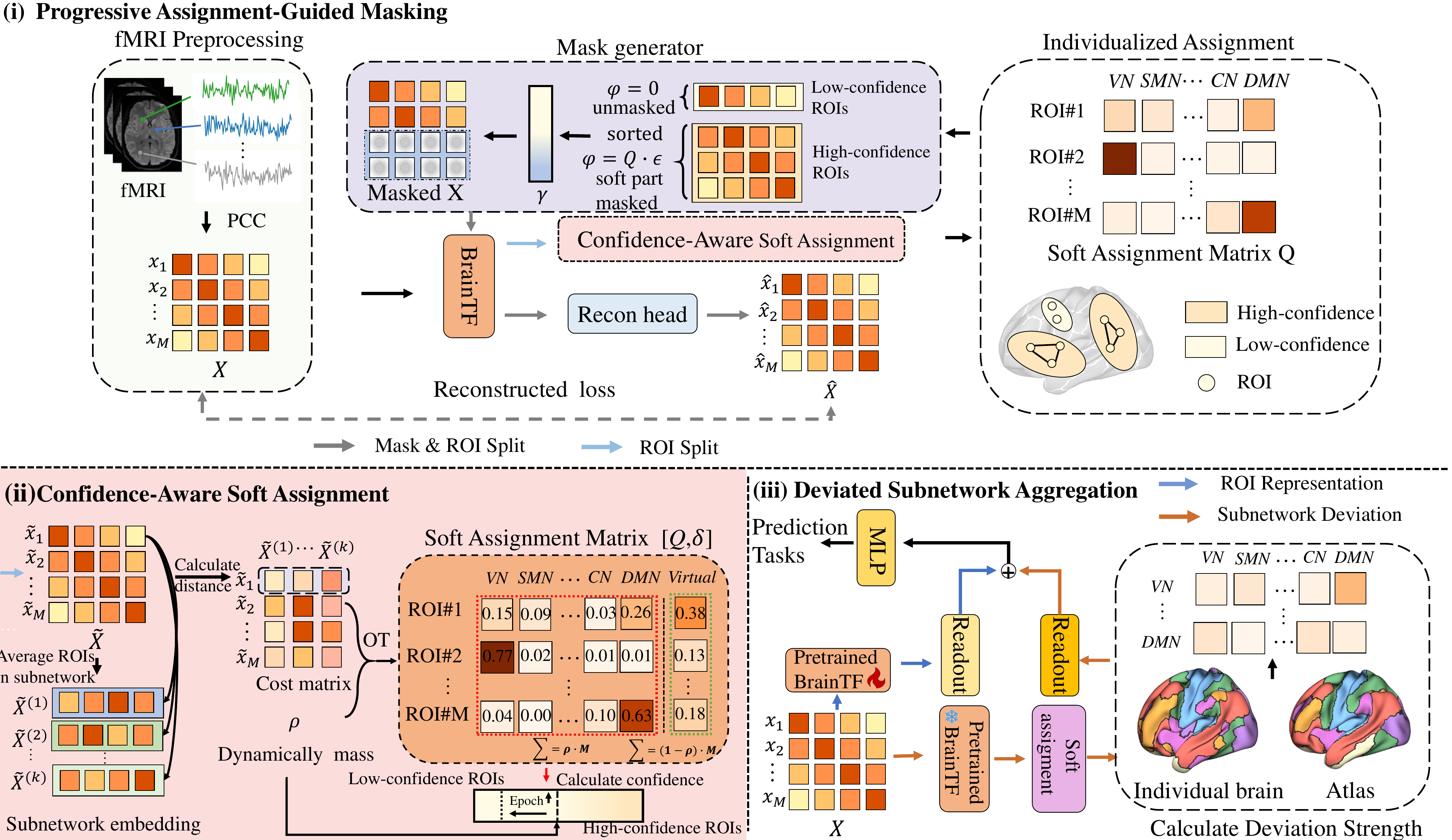}
\caption{Overview of BrainPICM. The framework consists of: (i) confidence-aware masking guided by soft assignment; (ii) progressive UOT-based soft assignment deriving ROI-level confidence via a virtual subnetwork; and (iii) deviation-aware subnetwork aggregation capturing functional reorganization for prediction. Recon: Reconstructed. PCC: Pearson Correlation Coefficient. $X$: Brain network. $x_i$: the i-th ROI with M features. $\hat{X}$: Reconstructed brain network. $\psi$: Masking probability. $\rho$: Dynamic transport mass. $\varepsilon$: Random subnetwork noise vector. $\gamma$: Binary mask vector.}
\label{fig2}
\end{figure*}

\section{Method}

\subsection{Overview}
As illustrated in Fig.~\ref{fig2}, the proposed BrainPICM is a progressive self-supervised framework that explicitly injects functional community structure into masked modeling of brain networks. In pre-training, the encoder learns ROI-wise representations and simultaneously estimates soft ROI-to-community affiliations with a progressive unbalanced optimal transport (UOT) solver. The resulting soft affiliations provide (i) soft ROI-to-community assignment  and (ii) ROI-level confidence scores, which together drive a curriculum-style masking policy: training starts from reliable community-consistent ROIs and gradually incorporates uncertain, potentially pathological ROIs. After pre-training, the encoder is frozen to provide stable community guidance. For downstream diagnosis, we further summarize how each subject deviates from a population-level community reference, and fuse ROI-level connectivity features with community-deviation signals in a lightweight dual-branch predictor. This design enables BrainPICM to learn representations that are both functionally consistent and sensitive to subject-specific heterogeneity.
% As illustrated in Fig.~\ref{fig2}, the proposed BrainPICM framework is structured around a progressive pipeline that links population-level priors with individualized brain representations. 
% During the pretraining phase, the model learns ROI-wise representations from functional connectivity matrices and constructs individualized subnetwork assignments using a progressive unbalanced optimal transport (UOT) solver. 
% These assignments produce soft ROI-to-subnetwork mappings and confidence scores that guide a curriculum-style masking process. 
% Subsequently, the pretrained encoder is frozen to provide stable assignment guidance. 
% The derived soft assignments are compared with atlas-based priors to generate a deviation matrix, which quantifies subnetwork-level reorganizations. 
% Finally, during fine-tuning, a dual-branch classifier is employed: one branch models the original ROI-level connectivity features, while the other processes the deviation matrix. 
% By integrating these two streams, BrainPICM captures both localized connectivity patterns and higher-level subnetwork deviations, resulting in more robust and interpretable predictions for brain disorder diagnosis.

\subsection{Problem Definition}
We represent the brain functional network as a symmetric matrix $X \in \mathbb{R}^{M \times M}$ based on an atlas with $M$ Regions of Interest (ROIs).
Each ROI is treated as a sequence element $x_i = X_{i,:} \in \mathbb{R}^{M}$, where the $M$-dimensional vector reflects its connectivity profile across the whole brain.
In this study, we use the Yeo ICN parcellation~\cite{yeo2011organization} to define $K=7$ canonical communities (DMN, VN, DAN, SAN, SMN, CN, LN), and construct a one-hot template assignment $Q_{\text{template}} \in \{0,1\}^{M \times K}$. 
Our goal is to learn a Transformer encoder $f_{\theta}$ via masked reconstruction while explicitly leveraging this community organization.
Specifically, we estimate soft ROI-to-community assignments $Q \in \mathbb{R}_+^{M \times K}$ and ROI-level confidence scores $q$, which guide curriculum-style masking; meanwhile, $Q$ is summarized into compact community-level deviation features for downstream diagnosis.

\subsection{Confidence-Aware Soft Assignment via Progressive UOT}
We obtain the soft ROI-to-community assignment matrix $Q \in \mathbb{R}_+^{M \times K}$ via a progressive unbalanced optimal transport (UOT) formulation. Given a brain network $X$, the  brain network Transformer (BrainTF) encoder \cite{kan2022brain,yang2024brainmass} extracts ROI-wise features, and we maintain an exponential moving average (EMA) feature bank $\tilde{X}$ for stable estimation. 
Based on the Yeo ICN parcellation, we compute the $k$-th community prototype on EMA features as $\tilde{x}^{(k)}=\frac{1}{|S_k|}\sum_{i\in S_k}\tilde{x}_i$, where $S_k$ denotes the set of ROIs assigned to community $k$ in the template. We then measure ROI--community affinity by cosine similarity $P_{i,k}=\cos(\tilde{x}_i,\tilde{x}^{(k)})$ and construct the transport cost matrix $C=-\log \mathrm{softmax}(P)$.
To accommodate uncertain ROIs that do not align well with the $K$ canonical communities, we introduce a virtual community to absorb the remaining assignment mass $\delta_i$. This allows the soft assignment $Q$ to be formulated as a progressive UOT problem, where $\nu$ is an atlas-derived prior distribution, $\lambda$ controls the strength of the prior regularization, and $\rho(t) \in (0,1]$ specifies the total mass transported to the $K$ real communities. With this formulation, $Q$ is obtained by the following objective:
\begin{equation}
    \min_{Q \in \Pi} \langle Q, C \rangle_F + \lambda \, \mathrm{KL} \left( Q^\top \mathbf{1}_M, \frac{\rho(t)}{K} \nu \right),
\end{equation}
\begin{equation}
    \text{s.t.} \quad \Pi = \left\{ Q \in \mathbb{R}_+^{M \times K} \mid Q\mathbf{1}_K \le \nu, \mathbf{1}_M^\top Q \mathbf{1}_K = \rho(t) \right\}.
\end{equation}
The remaining mass $1-\rho(t)$ is assigned to the virtual community, yielding an ROI-level confidence $q_i=1-\delta_i$. We increase $\rho(t)=\rho_0+(1-\rho_0)\exp\!\left(-5\left(1-\frac{t}{T}\right)^2\right)$ progressively, producing a curriculum-style transition from confident to uncertain ROIs and generating $Q$ and $q$ for subsequent masking. To solve this unbalanced OT problem, we adopt an entropy-regularized UOT solver based on the efficient scaling algorithm \cite{chizat2018scaling,zhang2024p}, where the soft assignment matrix is iteratively updated via Sinkhorn-style matrix scaling after introducing the virtual community.

\subsection{Progressive Assignment-Guided Masking}
Building upon the UOT-based assignments, we design a community-guided masking mechanism to inject functional community structure into masked reconstruction.
Given $X$, we sample a binary mask $\gamma \in \{0,1\}^M$ (with $\gamma_i=1$ indicating that ROI $i$ is masked) and form $X^{(t)}$ by replacing masked ROIs with a learnable mask token $x_{\text{mask}}\in\mathbb{R}^M$, followed by adding positional embeddings.
The encoder $f_\theta$ and a two-layer MLP head $g_\phi$ reconstruct the full network as:
\begin{equation}
\hat{X}=g_\phi\!\left(f_\theta\!\left(X^{(t)}\right)\right),
\end{equation}
optimized by a masked reconstruction loss:
\begin{equation}
\mathcal{L}_r=\frac{1}{\|\gamma\|_1}\sum_{i=1}^{M}\gamma_i\cdot\|\hat{X}_i-X_i\|^2.
\end{equation}

To decide which ROIs to mask at step $t$, we compute a curriculum-style masking score using the soft assignment vector $Q_i$ and confidence $q_i$:
\begin{equation}
\psi_i^{(t)}=
\begin{cases}
Q_i\cdot\epsilon, & \text{if } q_i \ge \mathrm{Quantile}_{1-\rho(t)}(q),\\
0, & \text{otherwise},
\end{cases}
\end{equation}
where $\epsilon\sim\mathcal{U}(0,1)^K$ introduces community-level stochasticity.
We then mask the top-$\lfloor \alpha M \rfloor$ ROIs with the largest $\psi^{(t)}$.
As $\rho(t)$ increases, the masking gradually expands from high-confidence ROIs to more uncertain regions, encouraging the encoder to learn community-consistent yet individualized representations.

\subsection{Deviated Subnetwork Aggregation}
To further account for heterogeneity in functional community organization, we compute an assignment deviation matrix $\Delta Q = Q - Q_{\text{template}}$, where $Q_{\text{template}}$ is the one-hot template assignment.
We then aggregate $\Delta Q$ into a flow matrix $D \in \mathbb{R}^{K \times K}$:
\begin{equation}
D[i, j] = \sum_{n \in S_i} \Delta Q[n, j], \quad i,j=1,\dots,K,
\end{equation}
where $S_i$ denotes the set of ROIs assigned to template community $i$, and we set $D[i,i]=0$.

For downstream disease diagnosis, the frozen encoder $f_\theta$ extracts global connectivity features $z_X$ via a readout operation (e.g., mean pooling over ROIs).
Simultaneously, the flow matrix $D$ is flattened and processed by an independent MLP to generate deviation representations $z_D$.
The final diagnostic prediction is:
\begin{equation}
\hat{y} = \mathrm{Softmax}\!\left(\mathrm{MLP}(z_X + z_D)\right).
\end{equation}
This dual-branch design couples fine-grained ROI connectivity with community-level deviation cues, enabling the predictor to jointly model local patterns and higher-order reorganization.

\section{Experiments}
\subsection{Experimental Setup}
\textbf{Datasets.} In this study, we conduct experiments on three real-world fMRI datasets, selected for their coverage of a spectrum of neurodevelopmental and neurodegenerative disorders and their widespread use. (a) Autism Brain Imaging Data Exchange (ABIDE-I) dataset \cite{di2014autism}.  A total of 1114 subjects are obtained in this dataset, comprising 586 Normal Controls (NC) and 528 subjects diagnosed with Autism spectrum disorder (ASD). (b) Attention Deficit Hyperactivity Disorder (ADHD-200) dataset \cite{adhd2012adhd}. This dataset includes 1235 subjects, comprising 711 NC and 524 patients diagnosed with ADHD. (c) Alzheimer’s Disease Neuroimaging Initiative (ADNI) dataset \cite{jack2008alzheimer}. Our analysis focused on 142 NC and 149 patients diagnosed with Alzheimer’s disease (AD). All fMRI images were preprocessed using the Configurable Pipeline for the Analysis of Connectomes (C-PAC) pipeline \cite{craddock2013towards}. This processing includes skull stripping, slice timing correction, motion correction, global mean intensity normalization, regression of nuisance signals using 24 motion parameters, band-pass filtering (0.01-0.08 Hz) and parcellated by Schaefer atlas \cite{schaefer2018local} into 100 ROIs. Pearson Correlation Coefficient was applied to measure functional connectivity.

\textbf{Baselines.} We compare our method with various baselines, including  supervised deep learning models (BrainNetCNN \cite{kawahara2017brainnetcnn}, BrainNetTF \cite{kan2022brain}, Com-BrainTF \cite{bannadabhavi2023community}, CAGT \cite{pei2025community}, BrainHGT \cite{ma2026brainhgt}, vanillaTF), and self-supervised learning frameworks (MAE \cite{he2022masked}, EvolvedMask \cite{feng2023evolved}, EAG-RS \cite{jung2023eag}, BrainMass \cite{yang2024brainmass}, TARDRL \cite{zhao2024tardrl}).

\textbf{Implementation Details.} In this study, all experiments were conducted on the PyTorch 2.4.1 platform using 4 NVIDIA 2080Ti GPUs with 11GB memory. For pretraining, we trained the model for 600 epochs using Adam optimizer with an initial learning rate of 3e-4 and a weight decay of 5e-5. For hyperparameters, we set the masking ratio $\alpha$ to 0.1 and $\lambda$ as 1. The initial value of $\rho$ is dataset-specific: 0.6 for ADNI, 0.8 for ABIDE, and 0.9 for ADHD. In this study, we determined these values by a grid search with a step of 0.1. In the downstream tasks, the model is trained for 50 epochs by using an early stopping strategy. To ensure fairness, we adopt a stratified sampling strategy that accounts for site-specific distributions when splitting the data into training (70\%), validation (10\%), and testing (20\%) sets, following established practice \cite{kan2022brain}. We evaluated both overall model performance and diagnostic relevance, reporting diagnosis accuracy (ACC) and area under the receiver operating characteristic curve (AUC), averaged over 10 random runs, with both mean and standard deviation on the test dataset.

\begin{table*}[t]
\centering
\setlength{\tabcolsep}{4pt}
\caption{Classification performance on ADHD-200, ABIDE-I and ADNI (Mean$\pm$Std). 
The best results are shown in bold and the second best are underlined.}
\label{tab:main}
\resizebox{\textwidth}{!}{
\begin{tabular}{llcccccc}
\toprule
\textbf{Category} & \textbf{Method} 
& \multicolumn{2}{c}{\textbf{ADHD-200}} 
& \multicolumn{2}{c}{\textbf{ABIDE-I}} 
& \multicolumn{2}{c}{\textbf{ADNI}} \\
\cmidrule(lr){3-4} \cmidrule(lr){5-6} \cmidrule(lr){7-8} 
& 
& ACC $\uparrow$ & AUC $\uparrow$ 
& ACC $\uparrow$ & AUC $\uparrow$ 
& ACC $\uparrow$ & AUC $\uparrow$ \\
\midrule

\multirow{8}{*}{Supervised}
& BrainNetCNN     & \underline{64.24$\pm$2.36} & 66.79$\pm$2.31 & 67.76$\pm$2.09 & 73.52$\pm$2.87 & 74.30$\pm$2.06 & 81.97$\pm$1.78 \\
& vanillaTF       & 62.65$\pm$2.51 & 66.19$\pm$2.46 & 67.51$\pm$2.27 & 72.58$\pm$2.27 & 74.57$\pm$6.91 & 81.72$\pm$3.94 \\
& BrainNetTF      & 63.86$\pm$2.71 & 67.77$\pm$2.32 & 70.03$\pm$1.96 & 74.38$\pm$2.17 & 77.12$\pm$4.10 & 83.50$\pm$4.20 \\
& Com-BrainTF     & 63.51$\pm$2.09 & 66.69$\pm$2.09 & 68.86$\pm$0.55 & 75.43$\pm$0.86 & 76.13$\pm$5.72 & 83.12$\pm$5.06 \\
& CAGT          & 62.93$\pm$2.78 & 66.95$\pm$3.53 & 68.27$\pm$2.42 & 74.04$\pm$2.69 & 76.61$\pm$4.02 & \underline{84.17$\pm$4.41} \\
& BrainHGT          & 63.47$\pm$0.88 & \underline{67.82$\pm$2.94} & 68.81$\pm$1.29 & 74.76$\pm$2.64 & \underline{78.98$\pm$4.10} & 83.96$\pm$2.84 \\
\midrule

\multirow{5}{*}{SSL}
& MAE             & 62.01$\pm$2.18 & 63.79$\pm$2.60 & 67.87$\pm$1.41 & 74.17$\pm$2.58 & 75.49$\pm$2.49 & 81.63$\pm$2.08 \\
& EvolvedMask     & 62.92$\pm$1.13 & 67.11$\pm$1.71 & 66.96$\pm$1.28 & 72.70$\pm$1.59 & 73.72$\pm$8.45 & 83.95$\pm$0.95 \\
& EAG-RS            & 62.97$\pm$0.78 & 66.47$\pm$2.98 & 69.90$\pm$3.46 & 74.51$\pm$3.05 & 73.56$\pm$1.65 & 79.60$\pm$2.87 \\
& TARDRL          & 61.37$\pm$1.37 & 61.33$\pm$1.67 & 68.36$\pm$0.19 & 72.45$\pm$1.13 & 74.41$\pm$3.51 & 80.86$\pm$4.32 \\
& BrainMass       & 64.12$\pm$2.08 & 66.41$\pm$1.78 & \underline{70.32$\pm$1.91} & \underline{76.21$\pm$1.13} & 76.78$\pm$2.74 & 81.22$\pm$3.01 \\
\midrule

Ours & \textbf{BrainPICM} 
& \textbf{67.37$\pm$0.83} 
& \textbf{68.37$\pm$1.31} 
& \textbf{72.03$\pm$0.82} 
& \textbf{77.06$\pm$1.25} 
& \textbf{82.03$\pm$2.75} 
& \textbf{84.33$\pm$3.62} \\

\bottomrule
\end{tabular}}
\end{table*}

\subsection{Results}
Table~\ref{tab:main} presents the averaged classification results (ACC and AUC) of our BrainPICM.
Our BrainPICM model achieves the best overall accuracy on all three datasets, with relative improvements of 1.71\%, 3.13\%, and 3.05\% over the second-best methods on ABIDE-I, ADHD-200, and ADNI, respectively. In addition to accuracy, BrainPICM also demonstrates consistently high AUC across the datasets. While some SSL methods do not show significant advantages over BrainNetTF, our method outperforms existing SSL baselines. This performance gain is attributed to the integration of individualized subnetwork assignment, a confidence-aware masking strategy, and deviation-aware feature aggregation, which together enable more precise and robust representation learning of subject-specific brain network structures.

\begin{figure}[t]
\centering
\includegraphics[width=1.0\columnwidth]{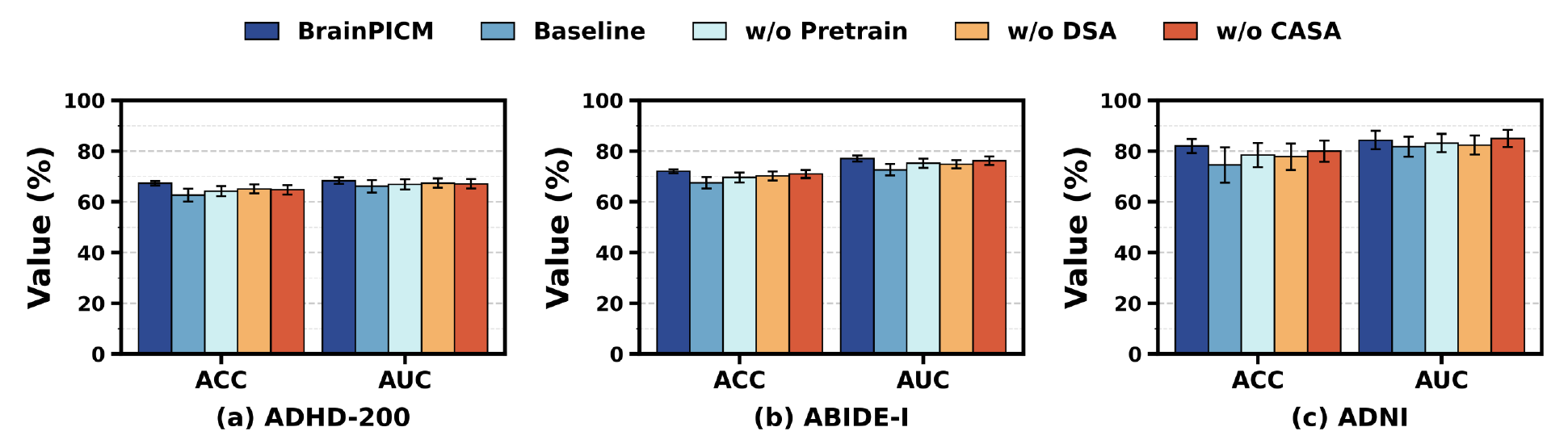} % Reduce the figure size so that it is slightly narrower than the column. Don't use precise values for figure width.This setup will avoid overfull boxes.
\caption{Ablation studies on the elements of Our BrainPICM.}
\label{fig3}
\end{figure}
%\textbf{Model Generalizability Analysis}

\subsection{Model Analysis}
\textbf{Ablation Studies.}
Figure~\ref{fig3} evaluates BrainPICM's core components: self-supervised pretraining, Confidence-Aware Soft Assignment (CASA), and Deviated Subnetwork Aggregation (DSA). Comparisons with a vanilla Transformer and a non-pretrained variant (using direct OT assignment) show that pretraining provides essential stability for individualized priors. Specifically, removing CASA increases assignment noise, while omitting DSA neglects critical group-to-individual reorganization signals. The consistent performance drop across all variants validates that these modules synergistically enhance representation learning.

% \begin{table}[t]
% \centering
% \scriptsizew1
% \setlength{\tabcolsep}{4pt}
% \begin{tabular}{l cc cc cc}
% \toprule
% \multirow{2}{*}{\textbf{Method}} & \multicolumn{2}{c}{\textbf{ADHD-200}} & \multicolumn{2}{c}{\textbf{ABIDE-I}} & \multicolumn{2}{c}{\textbf{ADNI}} \\
% \cmidrule(lr){2-3} \cmidrule(lr){4-5} \cmidrule(lr){6-7}
%  & \textbf{ACC$\uparrow$} & \textbf{AUC$\uparrow$} & \textbf{ACC$\uparrow$} & \textbf{AUC$\uparrow$} & \textbf{ACC$\uparrow$} & \textbf{AUC$\uparrow$} \\
% \midrule
% MAE + Ours         & 62.61 & 64.66 & 68.32 & 73.36 & 76.44 & 83.04 \\
% EvolvedMask + Ours & 63.39 & 65.99 & 67.46 & 72.21 & 74.92 & 78.85 \\
% HA + Ours          & 65.15 & 65.75 & 68.00 & 73.73 & 73.05 & 79.16 \\
% Ours               & \textbf{67.37} & \textbf{68.37} & \textbf{72.03} & \textbf{77.06} & \textbf{82.03} & \textbf{84.33} \\
% \bottomrule
% \end{tabular}
% \caption{Comparison of different masking strategies across three datasets. HA: Hard assignment.}
% \label{tab2}
% \end{table}

\begin{table}[t]
\centering
\scriptsize   
\setlength{\tabcolsep}{5pt}  
\begin{tabular}{c cc cc cc}
\toprule
\multirow{2}{*}{$\rho_0$} & \multicolumn{2}{c}{\textbf{Sigmoid}} & \multicolumn{2}{c}{\textbf{Linear}} & \multicolumn{2}{c}{\textbf{Fixed}} \\
\cmidrule(lr){2-3} \cmidrule(lr){4-5} \cmidrule(lr){6-7}
 & \textbf{ACC$\uparrow$} & \textbf{AUC$\uparrow$} & \textbf{ACC$\uparrow$} & \textbf{AUC$\uparrow$} & \textbf{ACC$\uparrow$} & \textbf{AUC$\uparrow$} \\
\midrule
0.6 & 82.03 & 84.33 & 78.14 & 83.58 & 75.42 & 81.17 \\
0.7 & 79.83 & 83.09 & 76.61 & 81.87 & 73.56 & 81.96 \\
0.8 & 81.53 & 84.01 & 79.15 & 83.43 & 75.25 & 82.09 \\
0.9 & 78.64 & 84.12 & 78.21 & 84.10 & 74.58 & 80.62 \\
\bottomrule
\end{tabular}
\caption{Analysis of $\rho_0$ with different ramp-up strategies on the ADNI dataset. 
Fixed means $\rho$ is held constant.}
\label{tab3}
\end{table}

% \textbf{The effects of mask strategy.}
% Table~\ref{tab2} analyzes the impact of different masking strategies on representation learning. The first two variants leverage subnetwork deviation signals derived from the input brain network, while “HA” applies hard assignment in place of soft assignment. The results highlight the effectiveness of our proposed soft assignment–guided masking, which captures finer-grained ROI-to-subnetwork correspondences. Moreover, the performance gains of MAE and EvolvedMask after integrating Deviated Subnetwork Aggregation further validate the utility of modeling subnetwork deviations.

\textbf{Analysis of $\rho$}
We compared fixed, linear, and sigmoid scheduling for the transport mass $\rho$. Table~\ref{tab3} shows that dynamic scheduling consistently outperforms fixed settings. The sigmoid ramp achieves the best results, confirming that progressively increasing transport mass effectively balances initial structural stability with late-stage individual refinement.

\textbf{Influence of Different Atlases}
We evaluated BrainPICM on the ABIDE-I dataset using different brain atlases (Schaefer 100 atlas, Schaefer 200 atlas, Craddock 200 atlas \cite{craddock2012whole}). As shown in Fig.~\ref{fig4}(a), BrainPICM achieves stable diagnostic accuracy across all atlases, demonstrating robustness to different parcellations.

\subsection{Biological Explanation}
\begin{figure}[t]
\centering
\includegraphics[width=0.9\columnwidth]{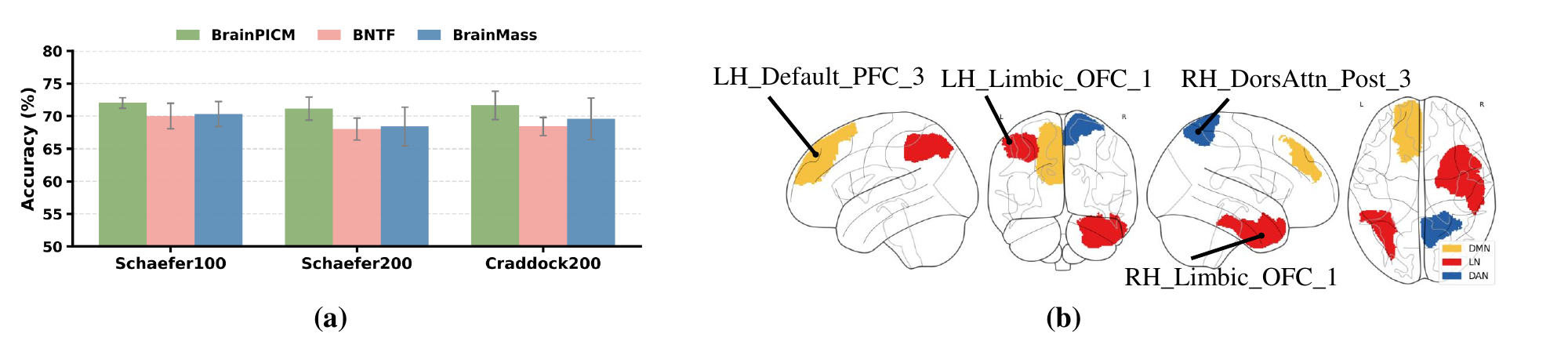} % Reduce the figure size so that it is slightly narrower than the column. Don't use precise values for figure width.This setup will avoid overfull boxes.
\caption{(a) Classification results on ABIDE-I dataset using different brain atlases; (b) Statistically significant ROIs ($p < 0.05$) identified by multivariate analysis.
LH: left hemisphere; RH: right hemisphere; PFC: prefrontal cortex; OFC: orbital frontal cortex. }
\label{fig4}
\end{figure}

Fig.~\ref{fig4}(b) illustrates the ROIs identified through multivariate analysis \cite{yang2021alteration}, indicating the significant involvement of the Default Mode Network (DMN), Dorsal Attention Network (DAN), and Limbic Network (LN). Notably, these systems are closely related to social cognition, attentional control, and affective processing, respectively, suggesting that the detected alterations may reflect core functional disruptions in autism. Consistent with these findings, prior studies have also reported connectivity abnormalities within these specific networks in individuals with autism \cite{he2018dynamic,hong2019atypical}.

\section{Conclusion}
In this paper, we introduced BrainPICM, a self-supervised framework that addresses the limitations of generic masking by injecting functional community structure into masked brain network modeling. By formulating ROI-to-community mapping as a progressive unbalanced optimal transport process, BrainPICM derives soft assignments and ROI-level confidence scores to drive curriculum-style community-guided masking from reliable to uncertain regions. We further summarize the soft assignments into compact community-level deviation features and fuse them with ROI-level connectivity features in a lightweight dual-branch predictor. Experiments on three fMRI datasets demonstrate that BrainPICM consistently outperforms state-of-the-art supervised and self-supervised baselines in diagnostic accuracy.

\begin{credits}
\subsubsection{\ackname}
This study is supported by grants from the National Key Research and Development Program of P.R.\ China (2025YFF0517803), the National Natural Science Foundation of P.R.\ China (62276081), Guangdong S\&T Programme (2025B0101130004), and the Shenzhen Science and Technology Program (CJGJZD\allowbreak 20230724093959002, JCYJ20250604145427037).
The funders only provided financial support for this study, and did not participate in the study design, data collection and analysis, manuscript writing, or submission decision-making. YY gratefully acknowledge support from the Alexander von Humboldt Foundation for his research fellowship.

\subsubsection{\discintname}
The authors declare no competing financial or personal interests that could have influenced this work.

\end{credits}

%
% ---- Bibliography ----
%
% BibTeX users should specify bibliography style 'splncs04'.
% References will then be sorted and formatted in the correct style.
%
\bibliographystyle{splncs04}
\bibliography{main}

@article{yang2023deep,
  title={A deep connectome learning network using graph convolution for connectome-disease association study},
  author={Yang, Yanwu and Ye, Chenfei and Ma, Ting},
  journal={Neural Networks},
  volume={164},
  pages={91--104},
  year={2023},
  publisher={Pergamon}
}

@inproceedings{yang2025hierarchical,
  title={Hierarchical Characterization of Brain Dynamics via State Space-Based Vector Quantization},
  author={Yang, Yanwu and Wolfers, Thomas},
  booktitle={International Conference on Medical Image Computing and Computer-Assisted Intervention},
  pages={394--404},
  year={2025},
  organization={Springer}
}

@article{keller2023personalized,
  title={Personalized functional brain network topography is associated with individual differences in youth cognition},
  author={Keller, Arielle S and Pines, Adam R and Shanmugan, Sheila and Sydnor, Valerie J and Cui, Zaixu and Bertolero, Maxwell A and Barzilay, Ran and Alexander-Bloch, Aaron F and Byington, Nora and Chen, Andrew and others},
  journal={Nature communications},
  volume={14},
  number={1},
  pages={8411},
  year={2023},
  publisher={Nature Publishing Group UK London}
}

@inproceedings{pei2025community,
  title={Community-aware graph transformer for brain disorder identification},
  author={Pei, Shengbing and Ma, Jiajun and Lv, Zhao and Zhang, Chao and Guan, Jihong},
  booktitle={Proceedings of the Thirty-Fourth International Joint Conference on Artificial Intelligence},
  pages={4191--4199},
  year={2025}
}

@article{kan2022brain,
  title={Brain network transformer},
  author={Kan, Xuan and Dai, Wei and Cui, Hejie and Zhang, Zilong and Guo, Ying and Yang, Carl},
  journal={Advances in Neural Information Processing Systems},
  volume={35},
  pages={25586--25599},
  year={2022}
}

@inproceedings{bannadabhavi2023community,
  title={Community-aware transformer for autism prediction in fmri connectome},
  author={Bannadabhavi, Anushree and Lee, Soojin and Deng, Wenlong and Ying, Rex and Li, Xiaoxiao},
  booktitle={International conference on medical image computing and computer-assisted intervention},
  pages={287--297},
  year={2023},
  organization={Springer}
}

@inproceedings{he2022masked,

  title={Masked autoencoders are scalable vision learners},

  author={He, Kaiming and Chen, Xinlei and Xie, Saining and Li, Yanghao and Doll{\'a}r, Piotr and Girshick, Ross},

  booktitle={Proceedings of the IEEE/CVF conference on computer vision and pattern recognition},

  pages={16000--16009},

  year={2022}

}

@inproceedings{feng2023evolved,

  title={Evolved part masking for self-supervised learning},

  author={Feng, Zhanzhou and Zhang, Shiliang},

  booktitle={Proceedings of the IEEE/CVF Conference on Computer Vision and Pattern Recognition},

  pages={10386--10395},

  year={2023}

}

@article{yang2024brainmass,

  title={Brainmass: Advancing brain network analysis for diagnosis with large-scale self-supervised learning},

  author={Yang, Yanwu and Ye, Chenfei and Su, Guinan and Zhang, Ziyao and Chang, Zhikai and Chen, Hairui and Chan, Piu and Yu, Yue and Ma, Ting},

  journal={IEEE transactions on medical imaging},

  volume={43},

  number={11},

  pages={4004--4016},

  year={2024},

  publisher={IEEE}

}

@article{segal2023regional,

  title={Regional, circuit and network heterogeneity of brain abnormalities in psychiatric disorders},

  author={Segal, Ashlea and Parkes, Linden and Aquino, Kevin and Kia, Seyed Mostafa and Wolfers, Thomas and Franke, Barbara and Hoogman, Martine and Beckmann, Christian F and Westlye, Lars T and Andreassen, Ole A and others},

  journal={Nature Neuroscience},

  volume={26},

  number={9},

  pages={1613--1629},

  year={2023},

  publisher={Nature Publishing Group US New York}

}

@inproceedings{yang2024advancing,

  title={Advancing Brain Imaging Analysis Step-by-Step via Progressive Self-paced Learning},

  author={Yang, Yanwu and Chen, Hairui and Hu, Jiesi and Guo, Xutao and Ma, Ting},

  booktitle={International Conference on Medical Image Computing and Computer-Assisted Intervention},

  pages={58--68},

  year={2024},

  organization={Springer}

}

@article{jung2023eag,
  title={EAG-RS: a novel explainability-guided ROI-selection framework for ASD Diagnosis via inter-regional relation learning},
  author={Jung, Wonsik and Jeon, Eunjin and Kang, Eunsong and Suk, Heung-Il},
  journal={IEEE Transactions on Medical Imaging},
  volume={43},
  number={4},
  pages={1400--1411},
  year={2023},
  publisher={IEEE}
}

@article{zhang2024p,
  title={{P}$^2${OT}: Progressive Partial Optimal Transport for Deep Imbalanced Clustering},
  author={Zhang, Chuyu and Ren, Hui and He, Xuming},
  journal={arXiv preprint arXiv:2401.09266},
  year={2024}
}

@article{adhd2012adhd,
  title={The ADHD-200 consortium: a model to advance the translational potential of neuroimaging in clinical neuroscience},
  author={{ADHD-200 Consortium}},
  journal={Frontiers in Systems Neuroscience},
  volume={6},
  pages={62},
  year={2012},
  publisher={Frontiers Research Foundation}
}

@article{di2014autism,
  title={The autism brain imaging data exchange: towards a large-scale evaluation of the intrinsic brain architecture in autism},
  author={Di Martino, Adriana and Yan, Chao-Gan and Li, Qingyang and Denio, Erin and Castellanos, Francisco X. and Alaerts, Kaat and Anderson, Jeffrey S. and Assaf, Michal and Bookheimer, Susan Y. and Dapretto, Mirella and others},
  journal={Molecular Psychiatry},
  volume={19},
  number={6},
  pages={659--667},
  year={2014},
  publisher={Nature Publishing Group}
}

@article{jack2008alzheimer,
  title={The Alzheimer's disease neuroimaging initiative (ADNI): MRI methods},
  author={Jack, Jr, Clifford R. and Bernstein, Matt A. and Fox, Nick C. and Thompson, Paul and Alexander, Gene and Harvey, Danielle and Borowski, Bret and Britson, Paula J. and Whitwell, Jennifer L. and Ward, Chadwick and others},
  journal={Journal of Magnetic Resonance Imaging},
  volume={27},
  number={4},
  pages={685--691},
  year={2008},
  publisher={Wiley Online Library}
}

@article{craddock2013towards,
  title={Towards automated analysis of connectomes: The configurable pipeline for the analysis of connectomes (C-PAC)},
  author={Craddock, Cameron and Sikka, Sharad and Cheung, Brian and Khanuja, Ranjeet and Ghosh, Satrajit S. and Yan, Chaogan and Li, Qingyang and Lurie, Daniel and Vogelstein, Joshua and Burns, Randal and others},
  journal={Frontiers in Neuroinformatics},
  volume={42},
  number={10.3389},
  year={2013}
}

@article{schaefer2018local,
  title={Local-global parcellation of the human cerebral cortex from intrinsic functional connectivity MRI},
  author={Schaefer, Alexander and Kong, Ru and Gordon, Evan M. and Laumann, Timothy O. and Zuo, Xi-Nian and Holmes, Avram J. and Eickhoff, Simon B. and Yeo, BT Thomas},
  journal={Cerebral Cortex},
  volume={28},
  number={9},
  pages={3095--3114},
  year={2018},
  publisher={Oxford University Press}
}

@article{kawahara2017brainnetcnn,
  title={BrainNetCNN: Convolutional neural networks for brain networks; towards predicting neurodevelopment},
  author={Kawahara, Jeremy and Brown, Colin J. and Miller, Steven P. and Booth, Brian G. and Chau, Vann and Grunau, Ruth E. and Zwicker, Jill G. and Hamarneh, Ghassan},
  journal={NeuroImage},
  volume={146},
  pages={1038--1049},
  year={2017},
  publisher={Elsevier}
}

@inproceedings{ma2026brainhgt,
  title={BrainHGT: A Hierarchical Graph Transformer for Interpretable Brain Network Analysis},
  author={Ma, Jiajun and Zhang, Yongchao and Zhang, Chao and Lv, Zhao and Pei, Shengbing},
  booktitle={Proceedings of the AAAI Conference on Artificial Intelligence},
  volume={40},
  number={21},
  pages={17617--17625},
  year={2026}
}

@inproceedings{zhao2024tardrl,
  title={TARDRL: Task-Aware Reconstruction for Dynamic Representation Learning of fMRI},
  author={Zhao, Yunxi and Nie, Dong and Chen, Geng and Wu, Xia and Zhang, Daoqiang and Wen, Xuyun},
  booktitle={International Conference on Medical Image Computing and Computer-Assisted Intervention},
  pages={700--710},
  year={2024},
  organization={Springer}
}

@article{dujardin2020tau,
  title={Tau molecular diversity contributes to clinical heterogeneity in Alzheimer’s disease},
  author={Dujardin, Simon and Commins, Caitlin and Lathuiliere, Aurelien and Beerepoot, Pieter and Fernandes, Analiese R and Kamath, Tarun V and De Los Santos, Mark B and Klickstein, Naomi and Corjuc, Diana L and Corjuc, Bianca T and others},
  journal={Nature medicine},
  volume={26},
  number={8},
  pages={1256--1263},
  year={2020},
  publisher={Nature Publishing Group US New York}
}

@article{sporns2016modular,
  title={Modular brain networks},
  author={Sporns, Olaf and Betzel, Richard F},
  journal={Annual review of psychology},
  volume={67},
  number={1},
  pages={613--640},
  year={2016},
  publisher={Annual Reviews}
}

@article{yeo2011organization,
  title={The organization of the human cerebral cortex estimated by intrinsic functional connectivity},
  author={Yeo, BT Thomas and Krienen, Fenna M and Sepulcre, Jorge and Sabuncu, Mert R and Lashkari, Danial and Hollinshead, Marisa and Roffman, Joshua L and Smoller, Jordan W and Z{\"o}llei, Lilla and Polimeni, Jonathan R and others},
  journal={Journal of neurophysiology},
  year={2011},
  publisher={American Physiological Society Bethesda, MD}
}

@article{yang2021alteration,
  title={Alteration of brain structural connectivity in progression of Parkinson's disease: a connectome-wide network analysis},
  author={Yang, Yanwu and Ye, Chenfei and Sun, Junyan and Liang, Li and Lv, Haiyan and Gao, Linlin and Fang, Jiliang and Ma, Ting and Wu, Tao},
  journal={NeuroImage: Clinical},
  volume={31},
  pages={102715},
  year={2021},
  publisher={Elsevier}
}

@article{he2018dynamic,
  title={Dynamic functional connectivity analysis reveals decreased variability of the default-mode network in developing autistic brain},
  author={He, Changchun and Chen, Yanchi and Jian, Taorong and Chen, Heng and Guo, Xiaonan and Wang, Jia and Wu, Lijie and Chen, Huafu and Duan, Xujun},
  journal={Autism Research},
  volume={11},
  number={11},
  pages={1479--1493},
  year={2018},
  publisher={Wiley Online Library}
}

@article{hong2019atypical,
  title={Atypical functional connectome hierarchy in autism},
  author={Hong, Seok-Jun and Vos de Wael, Reinder and Bethlehem, Richard AI and Lariviere, Sara and Paquola, Casey and Valk, Sofie L and Milham, Michael P and Di Martino, Adriana and Margulies, Daniel S and Smallwood, Jonathan and others},
  journal={Nature communications},
  volume={10},
  number={1},
  pages={1022},
  year={2019},
  publisher={Nature Publishing Group UK London}
}

@article{craddock2012whole,
  title={A whole brain fMRI atlas generated via spatially constrained spectral clustering},
  author={Craddock, R Cameron and James, G Andrew and Holtzheimer III, Paul E and Hu, Xiaoping P and Mayberg, Helen S},
  journal={Human brain mapping},
  volume={33},
  number={8},
  pages={1914--1928},
  year={2012},
  publisher={Wiley Online Library}
}

@article{chizat2018scaling,
  title={Scaling algorithms for unbalanced optimal transport problems},
  author={Chizat, Lenaic and Peyr{\'e}, Gabriel and Schmitzer, Bernhard and Vialard, Fran{\c{c}}ois-Xavier},
  journal={Mathematics of computation},
  volume={87},
  number={314},
  pages={2563--2609},
  year={2018}
}

\end{document}